\documentclass[10pt]{article}

\renewenvironment{abstract}{%
  \begin{center}
    \vspace*{2\baselineskip}
    {\sffamily\bfseries\large Abstract}
  \end{center}
}

\usepackage[margin=1in]{geometry}
\usepackage{float}
\usepackage{lmodern}
\usepackage{amsmath,amssymb}

\usepackage[round,authoryear]{natbib}

\usepackage{graphicx}
\usepackage{subcaption}
\usepackage{booktabs}
\usepackage{array}
\usepackage{multirow}

\usepackage{xcolor}
\usepackage{enumitem}
\setlist{nosep}

\definecolor{citeteal}{HTML}{2E7DB8}
\usepackage[colorlinks=true,citecolor=citeteal,linkcolor=citeteal,urlcolor=blue]{hyperref}

\usepackage{fancyhdr}
\fancypagestyle{plain}{%
  \fancyhf{}
  \setlength{\headsep}{8pt}

}

\usepackage{titlesec}
\titleformat{\section}{\normalfont\Large\bfseries}{\thesection}{1em}{}
\titleformat{\subsection}{\normalfont\large\bfseries}{\thesubsection}{1em}{}
\titleformat{\subsubsection}{\normalfont\bfseries}{\thesubsubsection}{1em}{}
\titleformat{\paragraph}[runin]{\normalfont\bfseries}{}{0pt}{}[.\ ]
\titlespacing*{\paragraph}{0pt}{1.5ex plus 1ex minus .2ex}{0.6em}

\definecolor{bestgreen}{HTML}{2E7D32}
\definecolor{worstterracotta}{HTML}{C1440E}
\newcommand{\best}[1]{\textcolor{bestgreen}{\textbf{#1}}}
\newcommand{\worst}[1]{\textcolor{worstterracotta}{\textbf{#1}}}
\newcommand{\figpanel}[2]{Fig.~\mbox{\ref{#1}#2}}
\definecolor{takeawaycolor}{RGB}{140,20,60}
\newcommand{\Takeaway}[1]{\noindent\textbf{\textcolor{takeawaycolor}{Takeaway.}} \textit{#1}}

\title{\Large\bfseries IRIS: A Visual Cortex-Inspired Framework for Analyzing Orientation Selectivity in Vision Transformers}

\author{%
\small
\begin{minipage}[t]{0.28\textwidth}
  \centering
  \textbf{Vaishnavi B Mohan} $^{*}$ \\ University of Washington, USA
\end{minipage}%
\hfill
\begin{minipage}[t]{0.24\textwidth}
  \centering
  \textbf{Vijayakrishna Naganoor} \\ Microsoft, USA
\end{minipage}%
\hfill
\begin{minipage}[t]{0.24\textwidth}
  \centering
  \textbf{Yashas Annadani} \\ TU Munich, Germany; Gladstone Institute, USA
\end{minipage}%
\hfill
\begin{minipage}[t]{0.2\textwidth}
  \centering
  \textbf{Shashank Hegde} \\ Nvidia, USA
\end{minipage}
\\[1.5em]
{\footnotesize $^{*}$Corresponding author: \texttt{vbmohan@uw.edu}}
}
\date{}

\author{%
\begin{minipage}[t]{0.40\textwidth}
  \centering
  \textbf{Vaishnavi B Mohan}$^{*}$ \\ \small University of Washington, USA
\end{minipage}%
\hfill
\begin{minipage}[t]{0.40\textwidth}
  \centering
  \textbf{Vijayakrishna Naganoor} \\ \small Microsoft, USA
\end{minipage}%
\\[1.5em]
\begin{minipage}[t]{0.40\textwidth}
  \centering
  \textbf{Yashas Annadani} \\ \small TU Munich, Germany
\end{minipage}%
\hfill
\begin{minipage}[t]{0.40\textwidth}
  \centering
  \textbf{Shashank Hegde} \\ \small Nvidia, USA
\end{minipage}%
\\[1.5em]
{\small $^{*}$Corresponding author: \texttt{vbmohan@uw.edu}}
}
\date{}

\begin{document}

\maketitle
\vspace{-3em}

\begin{abstract}
Vision transformers (ViTs) have become the \textit{de facto} standard for image encoding across many perception tasks. Despite their empirical success, it remains mechanistically unclear how they encode low-level features, given their lack of inductive biases: ViTs process information globally rather than relying on local structure. Biological visual systems, in contrast, build low-level features, such as orientation selectivity in the primary visual cortex, by combining information from small, localized regions of the visual field. These features are general-purpose representations, shared and required across multiple specialized neural pathways, unlike higher-level semantic features - which are task-specific. This raises the question of whether such biologically-grounded features arise in ViTs. In this work, we systematically study how orientation selectivity emerges in ViTs by introducing a suite of neuroscience-inspired metrics: representational similarity score (RSS), orientation recruitment score (ORS), and orientation tuning bandwidth to quantify how orientation is encoded in representational geometry and as a function of model depth.
Through extensive analysis, we find that: (1) the training paradigm is the strongest determinant of orientation selectivity, with models sharing an objective, peaking at comparable relative depths regardless of scale, (2) many units are orientation-selective early in training, with early-to-middle layers recruiting more such units over time, while deeper layers lose selectivity and broaden their tuning toward semantic encoding, and (3) our metrics offer a mechanistic heuristic for how many layers to unfreeze for best downstream generalization.
Our framework presents a way to track biologically-grounded features during ViT training, probes how desired properties are encoded in transformer representations, and builds a systematic understanding of how ViTs generalize across tasks.
\end{abstract}

\section{Introduction}
\label{sec:intro}

\begin{figure}[t]
    \centering
    \includegraphics[width=\linewidth]{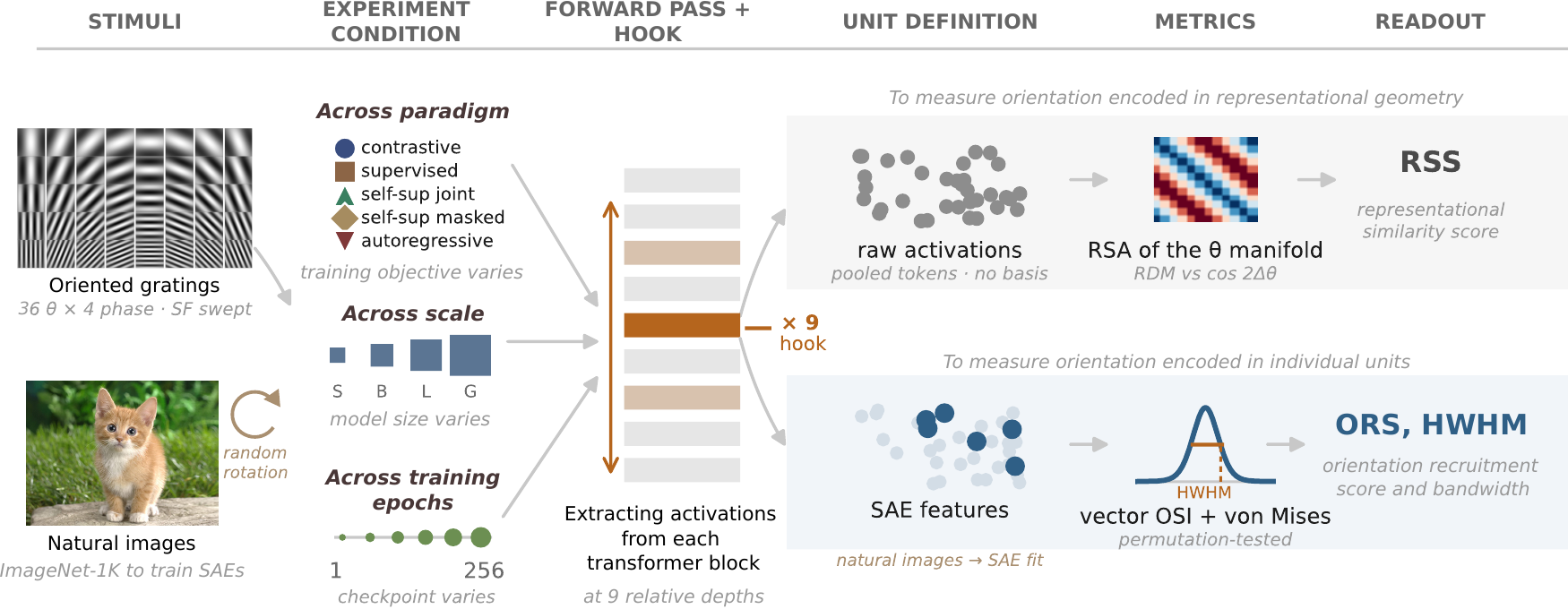}
    \caption{Probing pipeline. Oriented gratings and natural images pass
    through each frozen backbone, with activations recorded at 9 relative
    depths. We analyze these two ways: population-level similarity to the
    orientation manifold (RSS), and per-unit tuning from a sparse dictionary
    fit on natural images (vector OSI, ORS, von Mises bandwidth). We repeat
    this across three axes: pretraining paradigm, model scale, and training
    checkpoint.}
    \label{fig:pipeline}
\end{figure}

Vision Transformers (ViTs) \citep{dosovitskiy2021image} have become the dominant general-purpose backbone for computer vision tasks spanning recognition, segmentation \citep{li2023maskdino}, depth estimation \citep{lin2025depthanything3}, and embodied navigation \citep{Oquab2024}. Despite this empirical dominance, we understand comparatively little about how their internal representations are organized across depth, or how they learn those features over training.

This matters because a 2D image is an impoverished description of the world that produced it: a single object generates a vast, curved manifold of images under changes in viewpoint, illumination, and pose \citep{dicarlo2012does}, making stable, task-relevant structure an ill-posed inverse problem to recover \citep{marr1982vision}. A useful encoder must therefore be both rich, retaining low-level structure, and generalizable, so different downstream tasks can each read out what they need.

Biological vision solves this hierarchically: the primary visual cortex (V1) pools over small, local receptive fields, and many simple cells behave as oriented, bandpass filters described by Gabor functions \citep{hubel1962receptive}, a code shaped by the efficient-coding pressure of natural image statistics \citep{barlow1961possible, olshausen1996emergence} and recombined into increasingly complex, task-specific features \citep{hubel1962receptive, marr1982vision}. This code then relays to a dorsal stream equivariant to low-level geometry and a ventral stream invariant to it \citep{ungerleider1982two, dicarlo2012does}: one early code, two opposite downstream goals. CNNs make this locality explicit \citep{lecun1998gradient}; ViTs abandon it, mixing information globally from the first layer \citep{vaswani2017attention} and representing vision differently from CNNs \citep{raghu2021vision}, yet they are state of the art. Whether a V1-like oriented code emerges at all in such a globally mixing architecture is therefore a genuine empirical question, and a concrete, interpretable handle on general-purpose representation regardless of computational philosophy if it does.

Orientation is a canonical low-level feature with a precise tuning-curve characterization that emerges in the first layers of artificial vision models \citep{krizhevsky2012imagenet}, and it bears on a long-standing debate over whether a good encoder should be invariant or equivariant to input transformations \citep{cohen2016group}, usually treated as all-or-nothing. The brain suggests otherwise: invariance and equivariance coexist, resolved by region and task, equivariant in V1 and the dorsal stream, invariant in the ventral stream. Orientation separates the two regimes, since an equivariant representation preserves its geometry while an invariant one collapses it, and representational similarity analysis \citep{kriegeskorte2008representational} is a natural, basis-free measure of this equivariance, testing it layer by layer with a mechanistic lens rather than judging the network only by its output.

Our aim is twofold: through orientation selectivity, to identify where in a ViT this shared, generalizable code gives way to task-specific specialization, and, more broadly, to track how training shapes a network's internal representation rather than judging it only by its loss.

\paragraph{Overview of Contributions}
We present the Internal Representation Interpretability Suite (IRIS), with three contributions:
\begin{enumerate}
    \item A suite of neuroscience-inspired orientation metrics quantifying orientation selectivity across a ViT's layers, to be released as a python package --- described in Section~\ref{sec:method}.
    \item A systematic analysis of the emergence of orientation code based on training paradigm, model scale, and training dynamics, across five pretraining paradigms, five model families and all scale tiers --- described in Section~\ref{sec:experiments}.
    \item A mechanistically grounded criterion for selecting how many layers to unfreeze during fine-tuning to improve downstream generalization --- described in Section~\ref{sec:downstream}.
\end{enumerate}

\section{Related Work}
\label{sec:related}

\paragraph{Mechanistic interpretability of ViTs}
Recent progress in mechanistic interpretability has been driven in large part by tooling that makes it easy to open up a transformer and inspect the computation underlying its predictions, building on a framework for describing that computation as interpretable circuits \citep{elhage2021mathematical, olsson2022context}. Libraries such as TransformerLens \citep{nanda2022transformerlens}, NNsight \citep{fiottokaufman2025nnsight}, and, for vision and video, Prisma \citep{joseph2025prisma}, provide hooking interfaces for extracting raw activations and are now standard infrastructure in the field. A well-established technique is to project these activations into the sparse, overcomplete feature space of a sparse autoencoder (SAE), disentangling polysemantic activations into monosemantic features \citep{cunningham2023sparse, bricken2023monosemanticity}. First established for language models, these tools have since been adapted to vision transformers and CLIP: training patch-level SAEs for interpretable visual concepts \citep{hierarchical2025, rao2025steering}, and decomposing CLIP's representation via text-based analysis \citep{gandelsman2024interpreting, gandelsman2024decomposing}.

\paragraph{Aligning ViTs with human perception}
There have been different approaches in the past trying to evaluate similarities between deep learning computer vision models and biological vision. One of the initial works in this direction, Schrimpf et al. \citeyearpar{schrimpf2018brainscore}, introduced Brain-Score, based on how well a model's activations predict primate neural responses and how well its image-by-image error patterns match human behavior. In a related behavioral vein, Tuli et al. \citeyearpar{tuli2021convolutional} compare the error consistency of CNNs and Vision Transformers with humans. More recently, Raugel et al. \citeyearpar{raugel2025disentangling} compare model activations directly against fMRI and MEG brain recordings, finding that DINOv3's representations become brain-like in a developmental order that mirrors the maturation timeline of sensory versus prefrontal cortex. Other work instead measures model-human similarity through classical psychophysical measures, such as contrast sensitivity \citep{Akbarinia2023} and other low-level characteristics of the human visual system \citep{Cai_2025_CVPR}.

\section{Methods}
\label{sec:method}

We probe frozen, pretrained ViTs with a controlled battery of oriented stimuli: the stimulus battery (Sec.~\ref{sec:method:stim}), the probing protocol and units (Sec.~\ref{sec:method:probe}), and four complementary metrics of orientation encoding (Sec.~\ref{sec:method:metrics}).

\subsection{Stimulus Battery}
\label{sec:method:stim}

We probe each ViT model using a battery of synthetic sinusoidal gratings, a standard class of stimuli for characterizing orientation tuning in biological and, more recently, in vision foundation models \citep{hubel1962receptive,Cai_2025_CVPR}. Compared to natural images, these stimuli provide precise control over a small set of interpretable parameters, minimizing variation due to image content.
Each stimulus is defined as a sinusoidal grating modulated by a Gaussian envelope,
\begin{equation}
\label{eq:gabor}
\begin{aligned}
     x_\theta &= (x-c_x)\cos\theta + (y-c_y)\sin\theta \\
    y_\theta &= -(x-c_x)\sin\theta + (y-c_y)\cos\theta \\
G(x,y) &= c\,\cos\!\left(2\pi f\,x_\theta + \phi\right)\,
\exp\!\left(-\frac{x_\theta^{2}+y_\theta^{2}}{2\sigma^{2}}\right)
\end{aligned}
\end{equation}
where $(x,y)$ denote normalized image coordinates, $(c_x,c_y)$ is the center of the Gaussian envelope, and $(x_\theta,y_\theta)$ are the coordinates rotated by the grating orientation $\theta$. The stimulus consists of a sinusoidal carrier with spatial frequency $f$ and phase $\phi$, multiplied by a Gaussian envelope of width $\sigma$ and scaled by contrast $c$, which is fixed to $1$ throughout all experiments.

The battery samples orientation uniformly over $[0,\pi)$ at 36 evenly spaced values, phase from $\phi\in\{0,\frac{\pi}{2},\pi,\frac{3\pi}{2}\}$, and spatial frequency at five logarithmically spaced values (cycles per image), $f\in \left[1,0.75\frac{P}{2}\right]$, where $P$ is the model patch size and the upper bound is kept below the patch-wise Nyquist frequency to avoid aliasing.

\subsection{Probing and Units}
\label{sec:method:probe}

Each stimulus passes through a frozen ViT, with responses extracted from a chosen transformer block. Since models differ in depth, we compare responses at matched relative depths rather than fixed block indices: nine values $d \in \{0,0.1,0.25,0.4,0.5,0.6,0.75,0.9,1.0\}$, each mapped to a block of a model with $L$ blocks as $\mathrm{block}(d, L) = \mathrm{round}\!\big(d\,(L-1)\big)$, keeping comparisons consistent across architectures with different numbers of blocks.
Orientation selectivity, the degree to which a unit responds preferentially to one stimulus orientation over others, is the canonical example of feature selectivity in the visual system, epitomized by \emph{V1} simple cells \citep{hubel1962receptive}, and a natural first target for testing whether an artificial network builds an analogous, interpretable code. It is conventionally defined at the level of individual \emph{computational units}, a notion transformer representations do not straightforwardly admit, so we evaluate orientation tuning under three complementary readout bases that make progressively stronger assumptions about the underlying representation.

\paragraph{Raw coordinates}
We treat each (patch token, channel) coordinate of the residual stream as a unit ($D=d_{\text{model}}$ channels, one per patch). The residual stream has no privileged basis, so this is an assumption-light baseline for whether orientation aligns with the model's native coordinates.

\paragraph{MLP neurons}
We also treat each post-nonlinearity MLP neuron as a unit. Unlike the residual stream, which can be rotated arbitrarily without changing the network's function, the GELU nonlinearity fixes a privileged basis here, which we use to track a stable per-neuron identity across training checkpoints (Sec.~\ref{sec:results:training_epochs}).

\paragraph{SAE features}
We additionally project activations into the feature space learned by a sparse autoencoder (SAE), following standard practice in mechanistic interpretability \citep{cunningham2023sparse}. Each dictionary feature is treated as a unit, averaged over all patch tokens except the class token; this is our primary readout, since the learned features are the closest thing to a true monosemantic unit.

For all readout bases, responses are averaged over patch tokens to obtain a single response per stimulus. We combine responses across phases using the energy model,
\begin{equation}
\label{eq:energy}
\begin{aligned}
&\bar r_u(\theta, f) = \sqrt{\frac{1}{n_\phi}\sum_\phi r_u(\theta,\phi,f)^2} \\
    &\bar r_u(\theta) \equiv\bar r_u(\theta,f^\star_u), \qquad
f^\star_u = \arg\max_f \mathrm{OSI}\!\big(\bar r_u(\cdot,f)\big)
\end{aligned}
\end{equation}
where $r_u(\theta,\phi,f)$ is the response of unit $u$ to a stimulus with orientation $\theta$, phase $\phi$, and spatial frequency $f$, and $n_\phi$ is the number of sampled phases. For each unit, we then select the spatial frequency $f^\star_u$ that maximizes orientation selectivity (OSI, Sec.~\ref{sec:method:metrics}), yielding the phase-invariant tuning curve $\bar r_u(\theta)$ used in all subsequent unit-level analyses.
Unless otherwise stated, unit-level analyses are reported for the raw-coordinate and SAE-feature bases; the MLP-neuron basis is used only for the training-trajectory analysis (Sec.~\ref{sec:results:training_epochs}). Population-level analyses operate directly on the pooled activation vector $\mathbf{a}$.

\subsection{Orientation Metrics}
\label{sec:method:metrics}

This section defines four orientation metrics: the representational similarity score (RSS), our one population-level measure, and three unit-level measures, the Orientation Selectivity Index (OSI), the Orientation Recruitment Score (ORS), and tuning bandwidth, i.e.\ half-width at half-maximum (HWHM), the width of a unit's tuning curve at half its peak response above baseline.

\paragraph{Representational Similarity Score (RSS)}
We use representational similarity analysis (RSA), a standard neuroscience method for comparing representational geometry across systems with different coordinate bases \citep{kriegeskorte2008representational,Conwell2024}. For each stimulus $i$, we take the pooled raw activation (mean over patch tokens, class token excluded) as $\mathbf{a}_i$, $\ell_2$-normalize it, and correlate pairwise cosine dissimilarities with the dissimilarities predicted by stimulus orientation $\theta_i$ (circular distance, $180^\circ$ periodicity):
\begin{equation}
\centering
\begin{aligned}
&D^{\mathrm{act}}_{ij} = 1 - \frac{\mathbf{a}_i \cdot \mathbf{a}_j}
{\lVert \mathbf{a}_i\rVert\,\lVert \mathbf{a}_j\rVert}, \qquad
D^{\mathrm{orient}}_{ij} = \frac{\left|\angle\, e^{\,i\,2(\theta_i - \theta_j)}\right|}{\pi} \\
&\mathrm{RSS} = \mathrm{Pearson}\big(\{D^{\mathrm{act}}_{ij}\}_{i<j},\ \{D^{\mathrm{orient}}_{ij}\}_{i<j}\big)
\end{aligned}
\end{equation}
where $\angle$ denotes the argument of a complex number.

\textit{RSS is a population-level metric: a high value means stimuli of similar orientation are represented similarly and dissimilar ones are pushed apart, preserving orientation's circular geometry in the representation as a whole.} It depends only on pairwise similarities, so it is basis-invariant and directly comparable across models with different architectures, widths, and training objectives \citep{kriegeskorte2008representational}; we complement it with a basis-free linear decoding analysis (App.~\ref{app:decoding}) confirming orientation remains linearly decodable across depths even as this geometry changes.

\paragraph{Orientation Selectivity Index (OSI)}
For a single unit, we use the vector OSI, one minus the circular variance of its tuning curve. Since activations may be negative, we first rectify each curve, $\tilde r(\theta_k) = \bar r_u(\theta_k) - \min_j \bar r_u(\theta_j)$

\begin{equation}
\mathrm{OSI}_u = \frac{\left|\sum_k \tilde r_u(\theta_k)\,e^{\,i\,2\theta_k}\right|}
{\sum_k \tilde r_u(\theta_k)} \;\in[0,1].
\end{equation}
The harmonic $2\theta$ encodes orientation's $180^\circ$ periodicity. \textit{OSI is a continuous, per unit measure of how strongly a unit prefers a single orientation, using the whole curve rather than a hand picked pair of angles.}

\paragraph{Orientation Recruitment Score (ORS)}
To test whether a unit is genuinely orientation-selective rather than selective by chance, we build a null that mirrors the frequency-selection step in Eq.~\eqref{eq:energy}: for each of $B$ shuffles, we permute orientation labels on each of the five frequency-specific curves, recompute OSI at every frequency, and take the max across frequencies as that shuffle's null value. This gives the null the same five chances at a high OSI as the real statistic, so selecting the best frequency cannot alone inflate significance. A unit's $p$-value is the fraction of null values at least as large as its observed OSI; it counts as selective if $p<\alpha$, and ORS is the fraction of units passing this test.
\begin{equation}
\begin{aligned}
\mathrm{OSI}^{(b)}_{\mathrm{null}} &= \max_f \mathrm{OSI}\!\big(\pi_b(\bar r_u(\cdot,f))\big), \\
p_u &= \frac{1 + \sum_{b=1}^{B} \mathbf{1}\!\left[\,\mathrm{OSI}^{(b)}_{null} \ge \mathrm{OSI}_u\,\right]}{B+1}, \\
\mathrm{ORS} &= \frac{1}{N}\sum_{u} \mathbf{1}{\!\left[\,p_u < \alpha\,\right]},
\qquad \alpha = 0.05
\end{aligned}
\end{equation}
where $\pi_b$ permutes orientation labels for the $b$-th shuffle, $B$ is
the number of shuffles, $N$ the number of units, $\mathbf{1}[\cdot]$ the
indicator function, and $\alpha=0.05$.
\textit{ORS measures how much of the representation is allocated to orientation, a population count that the per-unit OSI does not provide.}

\paragraph{Tuning bandwidth (HWHM)}
OSI and ORS establish whether a unit is orientation-selective, and how many are, but not how sharply each is tuned. We therefore fit a von Mises (circular Gaussian) function to the tuning curve of each selective unit.
\begin{equation}
\begin{aligned}
   r(\theta) = b + A\,\exp\!\left(\kappa\left[\cos 2(\theta - \theta_{\mathrm{pref}}) - 1\right]\right), \\
    \mathrm{HWHM} = \tfrac{1}{2}\arccos\!\left(1 - \frac{\ln 2}{\kappa}\right)
\end{aligned}
\end{equation}
where $b$ and $A$ are the curve's floor and amplitude, $\theta_{\mathrm{pref}}$ is the unit's preferred orientation, and $\kappa$ is the concentration (inverse-variance) parameter fit to the curve. The closed form for HWHM is valid for $\kappa \ge \ln 2$; for the rare unit whose fit falls below that, no real half-max crossing exists, so we cap HWHM at $90^\circ$, half the $180^\circ$ period. \textit{Expressed in degrees, HWHM provides an architecture-independent measure of tuning precision that can be compared directly across models and with biological measurements of orientation tuning} \citep{Ringach2002}.

\section{Experimental Design}
\label{sec:experiments}

To understand what shapes the orientation code in ViTs, we ask how much of it is attributable to three factors, each isolated in its own study: the training objective (across paradigms), the model's size (across scale), and its training trajectory (across training epochs). Each study holds the other two factors fixed, and all three run the identical probe and stimulus battery described above.

\subsection{Across Paradigms}
\label{sec:results}
\begin{figure}[t]
  \centering
    \includegraphics[width=\textwidth]{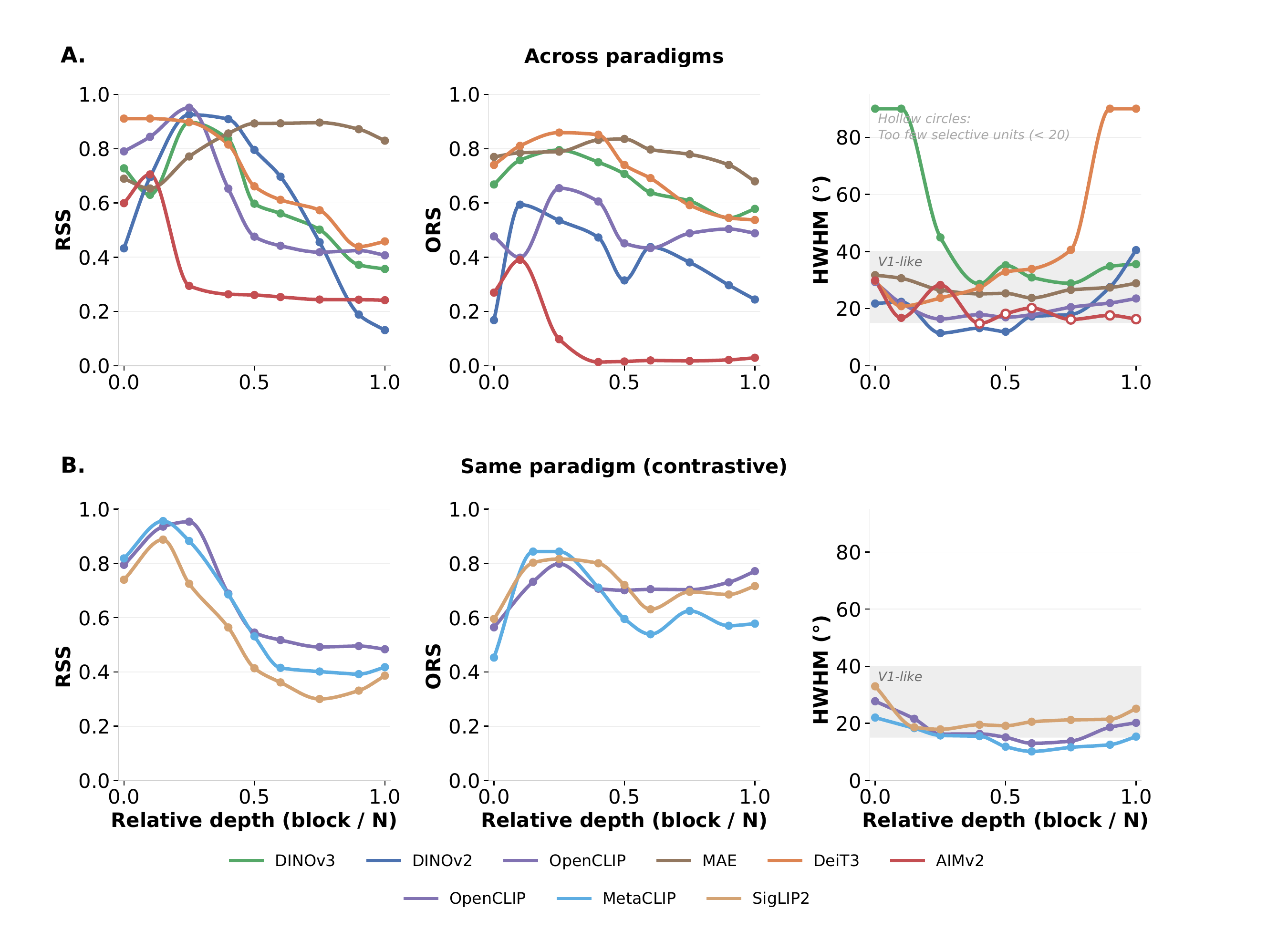}
    \caption{Orientation selectivity across training paradigms.
    \textbf{A.} We show RSS, ORS, and bandwidth (HWHM) against relative
    depth, one L-tier backbone per paradigm. All show an inverted-U RSS
    profile but diverge in where it decays: DINOv2/DINOv3 sustain it
    deepest, AIMv2 collapses earliest, and MAE keeps rising to the final
    layer. The shaded band marks V1-like bandwidths ($15^\circ$--$40^\circ$),
    and hollow circles mark blocks with fewer than 20 significant units.
    \textbf{B.} The same metrics within the contrastive family isolate
    data and loss as the only variables.}
  \label{fig:cross_model}
\end{figure}

\paragraph{Setup}
We study one representative ViT backbone from widely used pretraining paradigms: contrastive language supervision (OpenCLIP), supervised classification (DeiT III), self-distillation (DINOv2), masked autoencoding (MAE), and autoregressive prediction (AIMv2). We use one model per family at each scale tier, probed identically, so that any difference in the orientation code reflects the objective rather than architecture, scale, or the probe. In order to eliminate confounds like training data or loss function used, we add a within-contrastive control at the same scale: OpenCLIP (LAION-2B, softmax loss), MetaCLIP (curated CommonCrawl, softmax loss; data only changes), and SigLIP 2 (WebLI, sigmoid loss; data, loss, and patch size all change together). We use the contrastive family for this control because it's the only one with a clean, scale-matched set of checkpoints that vary data and loss one factor at a time; patch size, the main residual confound, is bounded by including both patch-14 and patch-16 models.

\paragraph{Observations}
RSS shows a common inverted-U profile across all five paradigms, but the DINO family sustains it out to a relative depth of 0.6 while AIMv2 collapses by 0.1, with MAE the outlier, still rising at the final layer. ORS shows the same early-layer bias in the number of selective units, with MAE retaining the most at depth, DINOv3 recruiting more than DINOv2, and AIMv2 dropping off earliest. Tuning bandwidth stays V1-like across models ($15^\circ$--$40^\circ$), though DINOv3's larger pool of selective units is more broadly tuned than DINOv2's smaller, sharper one.

\paragraph{Findings}
This inverted-U depth profile mirrors the primate visual hierarchy: weak selectivity in the retina, a peak in V1, and decline in higher, more semantic areas. DINOv2 and DINOv3 sustain this code over an extended depth range, consistent with their well-documented strength as frozen backbones \citep{Oquab2024,Simeoni2025}. MAE's own reported results match its RSS profile exactly: frozen or linear-probe MAE performs poorly (around 75\% top-1) while fine-tuned MAE is competitive (around 87.8\% for ViT-L) \citep{He2022}, exactly what an RSS that never hands off to a semantic code would predict, supporting our conjecture that MAE needs fine-tuning to convert that low-level code into task-relevant features. AIMv2-L14 is the sharpest contrast: billed by its authors as an encoder for general-purpose visual understanding \citep{Fini2025}, its RSS instead abandons low-level structure early, a profile that cannot plausibly support general-purpose visual features.

\Takeaway{Orientation selectivity reliably emerges in ViTs, sustained longest in the strongest frozen backbones. Training paradigm has the strongest effect, not data or architecture.}

\subsection{Across Scale}
\label{sec:results:scale}
\begin{figure}[t]
  \centering
    \includegraphics[width=\textwidth]{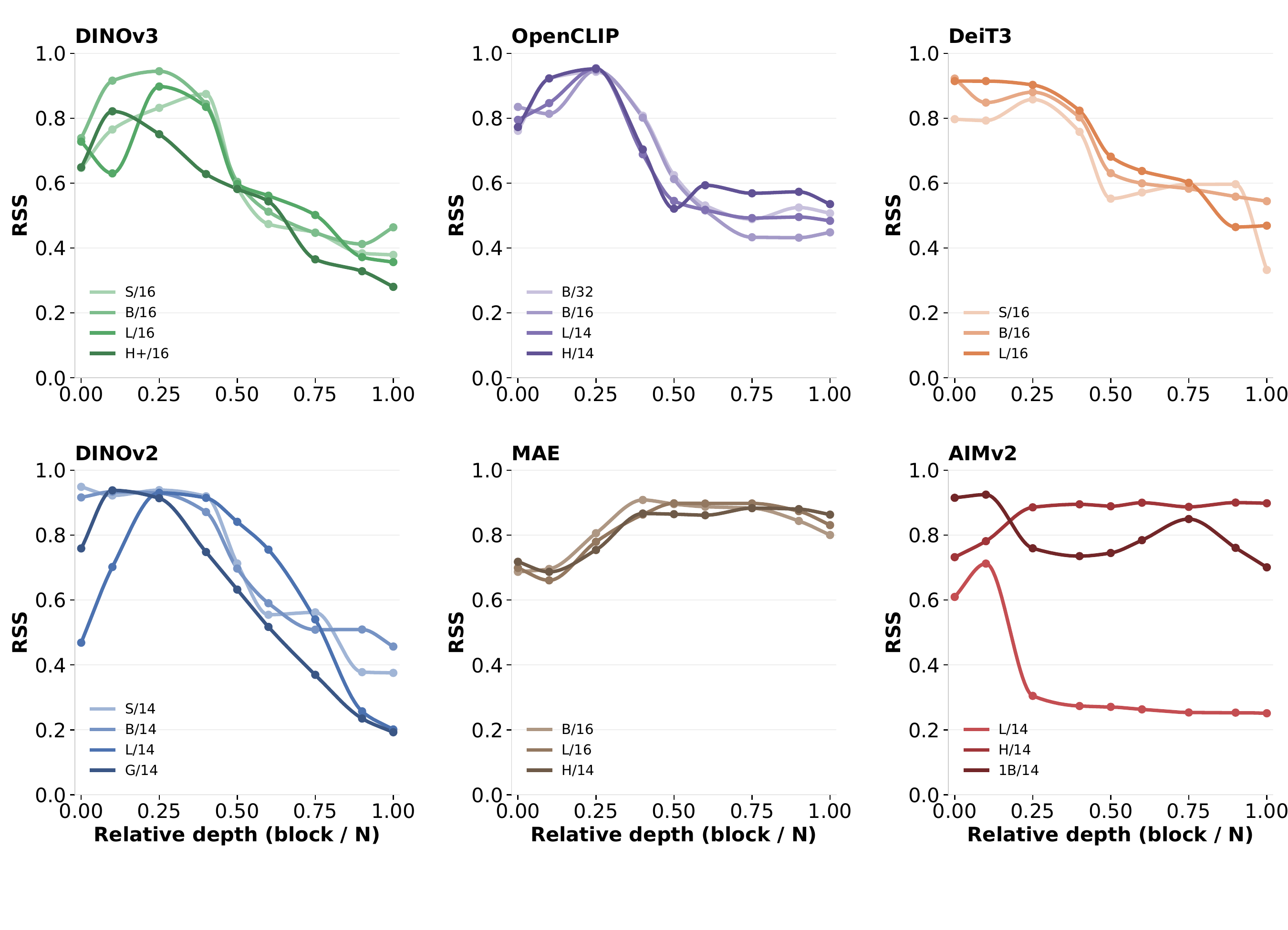}
    \caption{We plot RSS vs.\ relative depth, for each family, across tiers. Tiers overlay almost exactly within every family
    except AIMv2, the only one that fixes depth and scales width alone,
    whose RSS curve changes substantially with scale.}
  \label{fig:cross_scale}
\end{figure}
\paragraph{Setup}
Scale is a primary lever in modern vision models. We evaluate how the orientation code changes with scale by varying network size within family of models: OpenCLIP (B/32, B/16, L/14, H/14; LAION-2B), DINOv2 (S/B/L/G), DeiT III (S/B/L), MAE (B/L/H), and AIMv2 (L/H/1B), spanning roughly 22M to over 1B parameters. As before, we line models up by relative depth rather than by block number, since tiers within a family range from 12 to 48 blocks. Patch size only changes within OpenCLIP; the B/16 vs. B/32 pair, same size, different patch, solely isolates that effect.

\paragraph{Findings}
We use RSS for this comparison because it is comparable across models and layers without depending on a choice of basis. Across every family but one, the RSS curve is nearly identical across scale: the S, B, L, and G/H tiers overlap almost exactly when lined up by relative depth (\figpanel{fig:cross_scale}{}), indicating that relative depth, not block number, governs the strength of the orientation code. AIMv2 breaks this pattern: it is the only family that keeps 24 blocks at every tier and scales in width instead (App.~Table~\ref{tab:model_registry}), and its RSS curve is also the only one that changes drastically with scale.

This suggests depth and width play distinct roles: adding depth at constant width adds compositional degrees of freedom, so the feature space at a given relative depth keeps encoding the same code, while adding width instead changes what gets encoded there. This matches \citet{Greff2017} (added depth in residual networks refines existing features rather than replacing them) and \citet{Nguyen2021} (wide and deep networks do not learn the same things). AIMv2, the only family here that grows width while holding depth fixed, is the clearest case of the second effect.

\Takeaway{Adding more transformer blocks preserves how a feature is composed at each relative depth, while widening each block without depth changes what is encoded there. AIMv2, which scales only width, provides evidence for this.}

\subsection{Across Training Epochs}
\label{sec:results:training_epochs}

\begin{figure}[t]
  \centering
    \includegraphics[width=\textwidth]{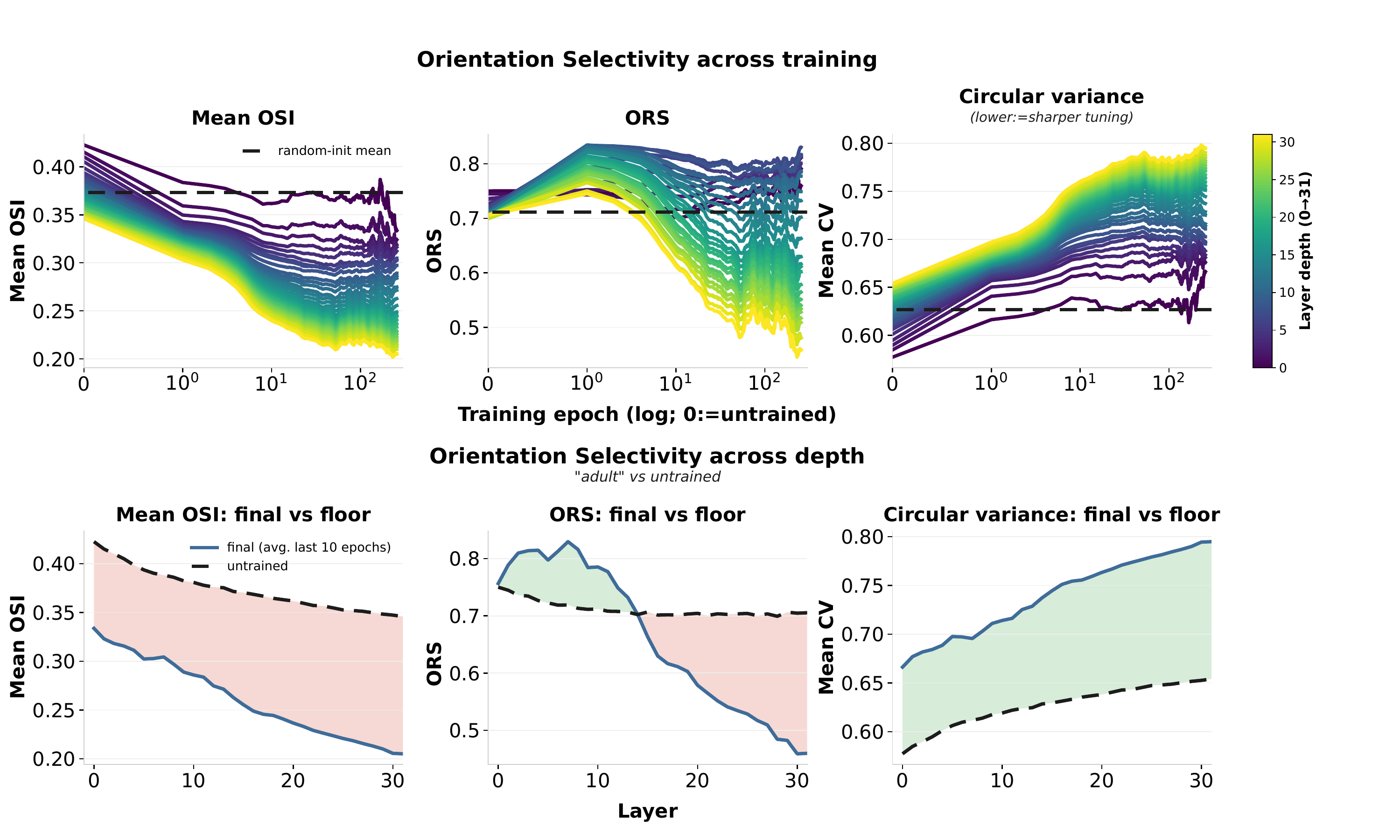}
    \caption{Orientation selectivity over OpenCLIP ViT-H/14's training
    trajectory. \emph{Left:} we plot mean OSI, ORS, and circular variance
    against training epoch (log scale, 0 = untrained), one line per layer
    (dark = shallow, light = deep), against the random-init floor (dashed).
    \emph{Right:} we plot the same metrics at the final checkpoint (last 10
    epochs, averaged) against the untrained floor, across layer depth;
    shading marks where training raises (green) or lowers (pink) a layer's
    value. Early-to-mid layers recruit more orientation-selective units
    than they start with; deep layers lose selectivity with training.}
  \label{fig:cross_training}
\end{figure}

While the comparisons above look at final trained models, here we ask how the orientation code forms over training by probing one network across its training checkpoints. In visual cortex, orientation-selective units are already present at eye-opening, and subsequent development sharpens individual neurons' tuning curves \citep{Chapman1993, White2001}. We ask the analogous question of a ViT: does orientation selectivity sharpen with training, and is that change gradual or concentrated in a critical-period-like window? Loss reports how well the network is doing, not what it is doing to get there; a biologically grounded, interpretable metric lets us see this directly, tells us when the representation matures enough to be useful downstream, and lets us compare its timeline against biological development.

\paragraph{Setup}
We track OpenCLIP ViT-H/14 across its public training checkpoints. At each checkpoint we probe the same oriented-grating battery used in the cross-model analysis, computing per-unit OSI, significance, and circular variance on raw MLP neurons, since the GELU nonlinearity fixes a privileged basis for neurons, unlike the residual stream, which can be rotated arbitrarily without changing the network's function. At each layer and epoch we report the mean OSI, the fraction of units passing significance (ORS), and the mean circular variance, compared against a random-init (untrained) floor computed the same way.

\paragraph{Findings}
We observe that earlier layers, which are the most orientation-selective, mature this feature very early in training, while later layers keep dropping it gradually across epochs (\figpanel{fig:cross_training}{}). This aligns with reports that deep networks converge bottom-up, with early layers reaching their final representation before later ones do \citep{Raghu2017}. As a conjecture, orientation maturity could serve as a proxy for when a layer is done learning, enabling progressive freezing schedules \citep{Brock2017} that pass gradients only to layers still developing their code. We have not validated that this preserves downstream performance, but if it holds, it could reduce training compute and give a mechanistic criterion for when to freeze each layer.

\Takeaway{Training reshapes orientation encoding rather than simply sharpening it: mid-layers recruit more orientation-selective units than an untrained network, while deeper layers trade orientation encoding for semantic content. This handoff matures earliest in the shallowest layers.}

\section{Implication for Downstream Tasks}
\label{sec:downstream}

If a ViT's early layers build a V1-like low-level code, general-purpose
and shared the way it is in visual cortex, that code should transfer well
across a wide range of otherwise unrelated downstream tasks. We test this by benchmarking the six scale-matched (L-tier) models
above as frozen encoders on 9 popular, divergently specialized tasks, using only a minimal
linear or $1{\times}1$-conv read-out so the decoder cannot compensate for
what the encoder lacks (full protocol and per-task results in
App.~\ref{sec:dt1}). Mean min--max normalized performance across these 9
tasks (App.~Eq.~\ref{eq:mean-norm-perf}) serves as a proxy for representation
generalizability, in the spirit of pooled-task transfer benchmarks
\citep{zhai2019vtab}. This pooled score tracks the shape of a model's RSS
curve, not merely its presence: DINO, whose RSS peaks broadly at mid-depth,
declining gradually, generalizes best; AIMv2-L, whose RSS collapses
earliest, generalizes worst; and MAE, whose orientation code never hands
off to semantic features and stays strong to the last layer, generalizes
almost as poorly despite the reverse profile.

This raises a sharper, practical question: if backbones differ in
\emph{where} their low-level code peaks, can that peak also tell us
where to start fine-tuning for best performance on a new task? Unfreeze too
little, only the deepest layers, and the model lacks the capacity to pack
in task-relevant information; unfreeze too much, and it discards low-level
structure the network has already built well. This choice matters most
where ViTs are reused as one shared trunk across many tasks, as in robotics
and autonomous driving \citep{majumdar2023vc1, hu2023uniad}.
Representations grow steadily more task-specific with depth
\citep{yosinski2014transferable}, and adapting only a subset of layers can
match or beat full fine-tuning depending on where the target task diverges
from pretraining \citep{lee2023surgical}. In practice, though,
the choice of which layer to start from is either an expensive sweep kept only
for its best result, or a fixed truncation depth used by convention,
especially in robotics \citep{nair2022r3m, majumdar2023vc1}. We test whether RSS, measured once on
the frozen backbone, can tell us where to start unfreezing without a
sweep.

\paragraph{Setup}
For a backbone with $L$
transformer blocks, we progressively unfreeze from block $k$ onward and
adapt blocks $[k:]$ with LoRA \citep{hu2022lora} ($r=8$, $\alpha=16$; full
hyperparameters and the exact depth grid in App.~\ref{app:lora-setup}),
sweeping $k$ over nine values spanning relative depth $d=k/(L-1)\in[0,1]$,
matched across backbones of different depth so results stay comparable. On
top of the backbone we train the same lightweight, task-specific decoder
heads as the frozen-backbone benchmark (App.~\ref{sec:dt1}). We fine-tune on
Taskonomy \citep{zamir2018taskonomy}, a popular public benchmark
spanning divergent visual tasks, using its tiny split since we sweep this
many depths, seeds, and backbones at once, across 9 visual downstream tasks
(segmentation, depth,
detection, surface normals, reshading, curvature, edges, 2D keypoints, and
Euclidean distance; full task definitions in App.~\ref{app:lora-setup}),
5 seeds per (task, $k$) (budget and split details in App.~\ref{app:lora-setup}), and
report mean min--max normalized performance across tasks for each depth
$k$ (App.~Eq.~\ref{eq:mean-norm-perf}).
We run this protocol across six backbones spanning family and scale.

\paragraph{RSS Predicts the Optimal Fine-Tuning Depth}
\begin{figure}[t]
  \centering
    \includegraphics[width=\linewidth]{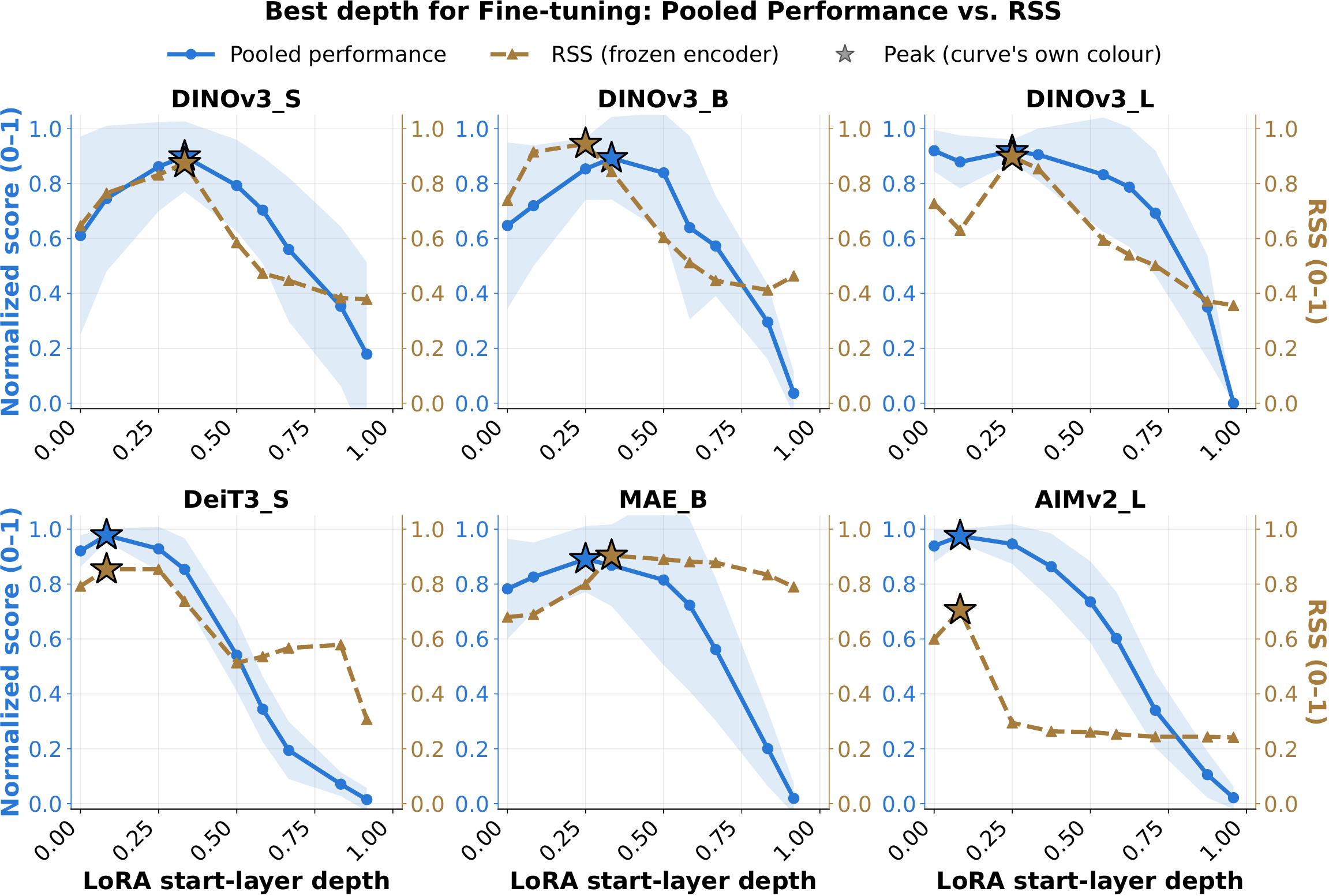}
    \caption{Pooled downstream performance (solid) against a-priori RSS on
    the frozen backbone (dashed), plotted against relative LoRA start-layer
    depth for six backbones spanning training paradigm. Performance is
    min--max normalized per backbone across tasks and seeds, following the same protocol as
    Figs.~\ref{fig:cross_model} and \ref{fig:cross_scale}. Stars mark each
    curve's own peak. The shaded band is the spread across seeds for pooled
    performance; RSS is computed once on the frozen backbone, so it has no
    seed spread.}
  \label{fig:fine-tuning}
\end{figure}
Once a backbone is chosen for a task using the frozen-backbone benchmark
(App.~\ref{sec:dt1}), RSS on that same frozen backbone tells us where to
start fine-tuning it. Figure~\ref{fig:fine-tuning} shows pooled performance when unfreezing and
fine-tuning from different depths, for each backbone. The depth that gives
the best pooled performance lines up with the depth where RSS, measured
once on the frozen encoder, peaks. The two peaks land on or almost exactly on top of each other across all
six backbones we test. For the DINOv3 family and for AIMv2-L/14, RSS's
peak lands exactly on the empirical optimum: both peak at relative depth
$d=0.25$ for DINOv3 (layer 4 for ViT-S/16, layer 6 for ViT-L/16), and at
the much earlier $d\approx0.1$ for AIMv2-L/14. This is not an artifact of
selecting a single favored metric:
scored against 9 other \textit{apriori} candidates computed from the same frozen
backbone (App.~\ref{app:apriori-ranking}), every one of them
performs poorly, and RSS is the only candidate whose peak reliably tracks
the true optimum.

The curves also make the cost of picking $k$ wrong clear. Performance falls
sharply as $k$ approaches the last few blocks: adapting only the deepest
layers leaves most of the network's capacity untouched, without enough room
to pack in task-relevant information. Faced with an unknown unfreeze depth, practitioners are left with two
options: sidestep the question and adapt the whole model with LoRA
\citep{hu2022lora, dettmers2023qlora}, i.e., $k=0$, or sweep across
unfreeze depths and keep whichever performs best. The former is often
suboptimal: our results show that unfreezing too early can discard
low-level structure the frozen backbone had already built well, hurting
rather than helping performance, especially for backbones like DINOv3 and MAE, which have already built a
rich low-level representation that generalizes well across a wide range of
tasks. The latter works, but is computationally expensive to repeat for
every new task, seed, and backbone. Based on our tests across six
backbones varying in scale and family, RSS is the best a-priori predictor
among those we compare, identifying the depth $k$ that gives the best mean
pooled performance from the frozen backbone alone, before any
fine-tuning.

\Takeaway{RSS, read off the frozen backbone before any fine-tuning,
predicts the best depth to unfreeze almost exactly, beating 9 other
a-priori candidates and the common default of adapting the whole model.}

\section{Limitations and Future Work}
\label{sec:discussion}
Our probing battery relies on synthetic sinusoidal gratings to isolate orientation tuning. While this provides precise parameter control, it does not capture the complexity of how these models process orientation in naturalistic images. Furthermore, our analysis is currently restricted to orientation selectivity; future work should expand this suite to include other foundational V1-like features, such as spatial frequency, color contrast, and motion direction.

\section{Conclusion}
\label{sec:conclusion}

We present IRIS, a suite of orientation metrics for probing any vision model and quantifying its orientation selectivity, a low-level feature central to visual encoding, to be released as a python package. IRIS offers a grey-box, mechanistic view of how orientation is encoded across the layers of a ViT: across the backbones we probe, training paradigm is the strongest determinant of how this depth-wise code emerges, relative depth rather than raw block count governs how it holds up with scale, and training reshapes the code rather than simply sharpening it, maturing earliest in the shallowest layers. Building on this, RSS's shape predicts a frozen encoder's downstream generalization, and that same curve's peak serves as an a-priori predictor of which layer to start fine-tuning from for the best downstream performance; the shape of its fall-off past that peak further predicts how much precision that choice requires, all read off the frozen backbone before any fine-tuning. Together, these results form a framework for three practical questions: which encoder to choose for a task, how to fine-tune a chosen encoder, and how to interpret an encoder's internal representations in terms grounded in visual neuroscience. We see this as a step toward a broader grey-box approach to vision encoders, an fMRI-like probe of the model itself, that could make studying and choosing among them more systematic.
\newpage
\bibliographystyle{plainnat}
\bibliography{references}

\newpage
\appendix
\section{Supplementary Materials}
\label{app:analysis}
\subsection{Feature Dictionary}
This section details the SAE readout basis introduced as `SAE features' in
Sec.~\ref{sec:method:probe}.

\paragraph{Why a dictionary} We use the sparse dictionary specifically for computing ORS (recruitment): determining which units are genuinely orientation-selective requires a unit basis that disentangles orientation from other, confounding attributes. Following standard practice in mechanistic interpretability \cite{cunningham2023sparse, bricken2023monosemanticity}, we learn an overcomplete sparse dictionary from each model's own activations and treat its individual features, rather than raw coordinates, as this per-unit basis.

\subsubsection{Training} We use a standard ReLU sparse autoencoder (SAE, Eq.~\ref{eq:sae}), trained with Adam (lr $10^{-3}$) to minimize reconstruction error plus an $\ell_1$ penalty on the code (Eq.~\ref{eq:sae-loss}), with decoder rows renormalized to unit norm every step so the sparsity penalty can't be evaded by rescaling. For every (model, block) pair, we independently fit one 2048-feature dictionary for 2000 steps, using minibatches of 1024 vectors and sparsity coefficient $\lambda=10^{-3}$. Each SAE is fit on that model's own activations at that block for 256 natural images from Imagenette \cite{Deng2009} rather than the gratings, so the dictionary matches the distribution the model actually computes with. Each image is pre-cropped oversized and rotated by an angle uniform on $[0^\circ,180^\circ)$ before a center crop, removing the population-level orientation bias natural photographs otherwise carry (the ``oblique effect''). This gives $256\times T$ training vectors, $T$ the number of tokens per image (patches plus class token). Dictionary size, $\lambda$, steps, and images are identical across every model, block, and seed, so recruitment, the selective fraction, is directly comparable across models and depths.

\begin{equation}\label{eq:sae}
\begin{aligned}
\mathbf{c}(\mathbf{x}) &= \mathrm{ReLU}\!\big(W_{\mathrm{enc}}(\mathbf{x}-\mathbf{b}_{\mathrm{dec}})+\mathbf{b}_{\mathrm{enc}}\big), \\[3pt]
\hat{\mathbf{x}} &= W_{\mathrm{dec}}\,\mathbf{c}(\mathbf{x}) + \mathbf{b}_{\mathrm{dec}} .
\end{aligned}
\end{equation}

where $\mathbf{x}$ is the block activation, $\mathbf{c}(\mathbf{x})$ the sparse code (feature activations) and $\hat{\mathbf{x}}$ its reconstruction, $W_{\mathrm{enc}},W_{\mathrm{dec}}$ and $\mathbf{b}_{\mathrm{enc}},\mathbf{b}_{\mathrm{dec}}$ the encoder/decoder weights and biases, $\lambda$ the sparsity coefficient, and $\mathcal{L}$ the reconstruction error. 

\begin{equation}\label{eq:sae-loss}
\mathcal{L} = \mathbb{E}\big[\lVert \hat{\mathbf{x}}-\mathbf{x}\rVert_2^2\big]
\;+\; \lambda\,\lVert \mathbf{c}(\mathbf{x})\rVert_1
\end{equation}

\begin{table}[t]
\centering
\resizebox{\textwidth}{!}{%
  \begin{tabular}{@{}l*{9}{c}@{}}
    \toprule
    \textbf{Backbone} &
    \rotatebox{75}{\shortstack[l]{Classfn\\{\scriptsize ImageNet-200}\\{\scriptsize top-1 acc.}}} &
    \rotatebox{75}{\shortstack[l]{Sem.\ segmtn\\{\scriptsize ADE20K}\\{\scriptsize mIoU}}} &
    \rotatebox{75}{\shortstack[l]{Depth\\{\scriptsize NYU Depth V2}\\{\scriptsize $\delta_1$}}} &
    \rotatebox{75}{\shortstack[l]{Pan.\ segmtn\\{\scriptsize COCO-Panoptic}\\{\scriptsize PQ (proxy mIoU)}}} &
    \rotatebox{75}{\shortstack[l]{Pose\\{\scriptsize COCO Keypts}\\{\scriptsize PCK@0.5}}} &
    \rotatebox{75}{\shortstack[l]{Detection\\{\scriptsize COCO val2017}\\{\scriptsize mAP@50}}} &
    \rotatebox{75}{\shortstack[l]{Retrieval\\{\scriptsize GLDv2-clean}\\{\scriptsize recall@1}}} &
    \rotatebox{75}{\shortstack[l]{Inst.\ segmtn\\{\scriptsize COCO val2017}\\{\scriptsize mask mAP@50}}} &
    \rotatebox{75}{\shortstack[l]{Surf. Nrml.\\{\scriptsize NYU-V2 (drvd.)}\\{\scriptsize acc.\ @11.25\textdegree}}} \\
    \midrule
    DINOv3  & 0.915 & \best{0.549} & \best{0.707} & \best{0.534} &  0.871         & \best{0.114} & 0.688         & \best{0.088} & 0.107         \\
    DINOv2  & 0.897        & 0.477        & 0.629        & 0.472        & \best{0.873}  & 0.085        & 0.531         & 0.060        & 0.095         \\
    SigLIP2 & 0.859        & 0.436        & 0.579        & 0.419        & 0.863         & 0.088        & \best{0.875}  & 0.062        & 0.087         \\
    MAE     & \worst{0.742}& \worst{0.309}& 0.586        & \worst{0.351}& 0.870         & 0.070        & \worst{0.375} & 0.056        & \best{0.118}  \\
    DeiT3   & \best{0.919} & 0.389        & 0.547        & 0.437        & 0.848         & 0.051        & 0.531         & 0.039        & \worst{0.083} \\
    AIMv2   & 0.883        & 0.342        & \worst{0.491} & 0.365        & \worst{0.790} & \worst{0.040}& \best{0.875}  & \worst{0.030}& 0.084         \\
    \bottomrule
  \end{tabular}%
  }
  \caption{Per-task validation performance for each frozen, L-tier backbone
  (Table~\ref{tab:model_registry}) across all 9 benchmark tasks (dataset and
  decoder head for each task in Table~\ref{tab:benchmark}; full protocol in
  App.~\ref{sec:dt1}). Each entry is the best checkpoint's score on that
  task's own validation split and primary metric. Green indicates the
  best-performing backbone on a task and terracotta the worst; retrieval is
  a tie between two backbones. Metrics differ in scale across columns
  (accuracy, mAP, mIoU, \ldots) and are not directly comparable across
  tasks; Fig.~\ref{fig:benchmark} and Eq.~\ref{eq:mean-norm-perf} report the
  pooled comparison used elsewhere in the paper.}
  \label{tab:benchmark_results}
\end{table}

\subsection{Orientation Decoding} \label{app:decoding}
We test whether orientation is linearly recoverable from the population by performing ridge regression from the pooled raw activation $\mathbf{a}_i$ (mean over patch tokens, class token excluded) to orientation. Since orientation is $180^\circ$-periodic, the regression target is $(\sin 2\theta_i, \cos 2\theta_i)$ rather than $\theta_i$; we use 5-fold cross-validation within the grating battery so no stimulus predicts itself:
\begin{equation}\label{eq:decode-target}
\mathbf{y}_i = (\sin 2\theta_i,\ \cos 2\theta_i),
\qquad
\hat{\mathbf{y}}_i = \mathrm{Ridge}_{\mathrm{CV}}(\mathbf{a}_i).
\end{equation}
Decodability is the cross-validated coefficient of determination over both targets (1 = perfect recovery, 0 = mean-predictor baseline); recovering a predicted orientation from $\hat{\mathbf{y}}_i$ also gives a mean absolute circular error in degrees:
\begin{equation}\label{eq:decodability}
\begin{aligned}
\mathrm{decodability} &= 1 - \frac{\sum_i \lVert \mathbf{y}_i - \hat{\mathbf{y}}_i\rVert^2}
{\sum_i \lVert \mathbf{y}_i - \bar{\mathbf{y}}\rVert^2}, \\[4pt]
\hat\theta_i &= \tfrac{1}{2}\operatorname{atan2}(\hat y_{i,\sin},\, \hat y_{i,\cos}).
\end{aligned}
\end{equation}
Because same-orientation stimuli of different phase and spatial frequency fall on both sides of the cross-validation splits, the readout must generalize across those nuisances rather than memorize a grating.

We find decodability exceeds 95\% across every model and layer we probe, even in layers where RSS has already declined. RSS and decodability measure different things: RSS asks whether orientation dominates the representation's overall similarity structure, while decodability only asks whether some linear direction still tracks orientation, dominant or not. Their dissociation shows that a low RSS does not mean a layer has erased orientation -- the information is still there, linearly readable; it has just stopped being the axis that organizes the layer's geometry, likely because other, more task-relevant features have taken over that role.

\subsection{RSS dictates generalizability} \label{sec:dt1}

\paragraph{Context} V1 encodes orientation. It is the shared early stage
of the visual system, which then diverges into two specialized pathways:
dorsal, for spatial localization, and ventral, for recognition
\citep{ungerleider1982two}. This suggests that a model building this
low-level code early can generalize broadly across tasks.

Testing this requires many tasks, not one: strong performance on a single
task may just reflect a representation suited to that task. Only
performance that holds across tasks with no shared structure -- semantic
versus geometric, whole-image versus dense-pixel -- is evidence of a
genuinely general-purpose code. This is why we benchmark across 9 popular,
divergently specialized visual tasks (Table~\ref{tab:benchmark}) rather
than a single one.

We keep every backbone frozen throughout and train only a lightweight,
task-specific decoder head on top of it. Fine-tuning the backbone itself
would let each task-specific head reshape the encoder, confounding any
test of whether the pretrained representation is general-purpose.

\begin{figure}[ht]
  \centering
    \includegraphics[width=0.6\textwidth]{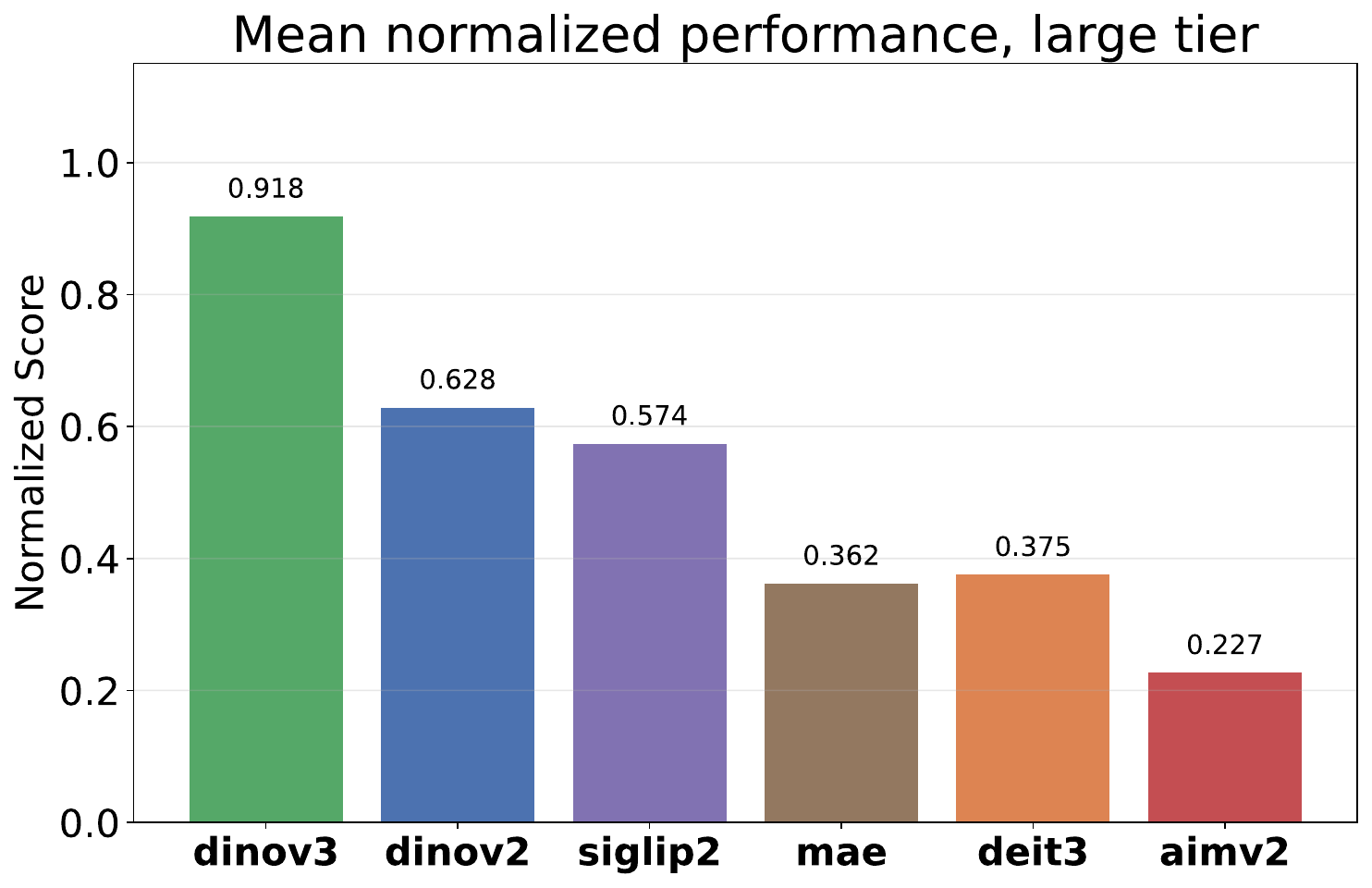}
    \caption{Mean min--max normalized performance $\overline{S}(b)$
    (Eq.~\ref{eq:mean-norm-perf}) for the six L-tier frozen backbones,
    pooled across all 9 tasks: within each task the weakest backbone
    scores 0 and the strongest 1, and $\overline{S}(b)$ averages this
    per-task rank across tasks -- a relative ranking, not an absolute
    score. DINOv3 generalizes best ($\overline{S}=0.918$), followed by
    DINOv2 (0.628) and SigLIP2 (0.574); MAE (0.362) and DeiT3 (0.375) are
    mid-low, and AIMv2 is worst (0.227). This ordering is specific to
    these 9 tasks at the L tier and may shift with task selection or
    model scale.}
  \label{fig:benchmark}
\end{figure}

This appendix reports that frozen-backbone benchmark; the main text uses
only its pooled score (Fig.~\ref{fig:benchmark}) and its role as the
starting point for the fine-tuning-depth sweep (App.~\ref{app:lora-setup}).

\paragraph{Setup} Per-task numbers for individual backbones exist
scattered across prior work, but no single study evaluates this set of
tasks and models under one consistent protocol; differing splits,
decoders, and training budgets make those numbers incomparable across
papers. We therefore run this benchmark ourselves, for a controlled,
apples-to-apples comparison: the six L-tier models
(Table~\ref{tab:model_registry}) on 9 tasks (Table~\ref{tab:benchmark}).
Each frozen encoder is paired with a minimal, task-specific linear head:
global pooling + linear for whole-image tasks (classification,
retrieval; retrieval additionally $\ell_2$-normalized), or one or more
parallel $1\times1$ convolutions for dense tasks -- one for
segmentation/depth/normals, CenterNet-style heatmap/box/offset branches
(plus a mask branch for instance segmentation) for detection, and a
$K$-channel conv for pose. No head has a hidden layer, so it cannot
compensate for what the encoder lacks. For each (backbone, task) pair, we
train a separate decoder independently -- its own dataset, head, loss, and
metric -- closer to
standard linear-probe suites (e.g.\ those reporting DINOv2/DINOv3 backbone
quality) than to a joint multi-task model. The backbone stays entirely
frozen; we use AdamW (weight decay $10^{-4}$, learning rate $3\times10^{-4}$,
no schedule), batch size 64, and each backbone's own native resolution:
$224\times224$ for every model except SigLIP2, run at its native
$256\times256$.

\paragraph{Datasets} We chose these 9 tasks because they are among the
most commonly used tasks for benchmarking visual encoders, and varied
enough from one another to test the generalizability claim: whole-image
classification and retrieval, dense per-pixel segmentation/depth/normals,
and instance-level detection and pose. Table~\ref{tab:benchmark} lists the
9 tasks, datasets, and metrics; each dataset is the most established
benchmark for its task: classification (Tiny-ImageNet-200),
detection/instance segmentation/pose (COCO val2017), semantic segmentation
(ADE20K), panoptic segmentation (COCO-Panoptic), depth/surface normals
(NYU Depth V2), and retrieval (a GLDv2 mini-subset). Panoptic segmentation
is scored by per-pixel mIoU over the combined category map, surface
normals are derived from each depth map via local-plane fitting, and
retrieval is scored by in-batch recall@1 -- standard, lightweight
protocols for benchmarking representation quality at this scale.

\paragraph{Training and evaluation} For each (backbone, task) pair, we
freeze the ViT backbone and train only the task-specific decoder head
end-to-end. Classification and semantic segmentation use each dataset's
own official split; the rest use a deterministic 90:10 split by sorted
index, fixed across backbones. Epoch count is computed automatically per
task from its own training-set size, targeting a comparable optimization
budget rather than a fixed epoch count; dataset size and epoch budget for
each task are given in Table~\ref{tab:benchmark}. We evaluate roughly 10
times per run regardless of its length and keep the best checkpoint on the
primary metric -- best-checkpoint selection over a fixed budget, not early
stopping.

\paragraph{Pooling into one score} To combine 9 differently-scaled
metrics, let $s(t,b)$ be backbone $b$'s best score on task $t$
(Table~\ref{tab:benchmark_results}); we min--max normalize each task across
the six backbones,
\begin{align}
  \hat{s}(t,b) &= \frac{s(t,b) - \min_{b'} s(t,b')}{\max_{b'} s(t,b') - \min_{b'} s(t,b')} \in [0,1] \notag \\
  \overline{S}(b) &= \frac{1}{|T_b|}\sum_{t \in T_b} \hat{s}(t,b) \label{eq:mean-norm-perf}
\end{align}
where $b'$ ranges over all 6 backbones with a result on $t$, so
$\hat{s}(t,b)\in[0,1]$ (0 = weakest, 1 = strongest on $t$) regardless of raw
scale, and $\overline{S}(b)$ is the mean over $T_b$ (all 9 tasks here), the
number plotted in \figpanel{fig:benchmark}. This pooling is deliberately
\emph{global}: $b'$ ranges over all six backbones, so $\overline{S}(b)$
answers which backbone is best overall. The fine-tuning-depth sweep
(App.~\ref{app:lora-setup}) reuses this equation for a different question,
how performance changes with unfreeze depth \emph{within} one chosen
backbone, so there $b'$ ranges only over that backbone's own depths,
normalized independently per backbone rather than shared across backbones.

\paragraph{Results} Table~\ref{tab:benchmark_results} gives every
(backbone, task) score; Fig.~\ref{fig:benchmark} pools them into
$\overline{S}(b)$. DINOv3 generalizes best ($\overline{S}=0.918$), then
DINOv2 ($0.628$) and SigLIP2 ($0.574$); MAE ($0.362$) and DeiT3 ($0.375$)
are similarly mid-low despite very different training objectives, and
AIMv2 is worst ($0.227$) -- matching what \figpanel{fig:cross_model}
predicts from RSS alone: DINOv2/DINOv3 sustain orientation selectivity
deepest and generalize best, AIMv2 collapses earliest and generalizes
worst, consistent with the V1-to-dorsal/ventral story above.

\begin{table}[t]
\centering
\resizebox{\textwidth}{!}{%
\small
\setlength{\tabcolsep}{4pt}
\begin{tabular}{l ccccccccc}
\toprule
\textbf{Tier} & \textbf{OpenCLIP} & \textbf{DeiT III} & \textbf{DINOv2} & \textbf{DINOv3} & \textbf{MAE} & \textbf{AIMv2} & \textbf{MetaCLIP} & \textbf{SigLIP2} \\
& \footnotesize{contrastive} & \footnotesize{supervised} & \footnotesize{self-sup, joint} & \footnotesize{self-sup, joint} & \footnotesize{self-sup, masked} & \footnotesize{autoregressive} & \footnotesize{contrastive} & \footnotesize{contrastive} \\
\midrule
Small       & --        & 22M / 12  & 22M / 12  & 21M / 12  & --        & --        & --        & --        \\
Base        & 87M / 12  & 87M / 12  & 86M / 12  & 86M / 12  & 87M / 12  & --        & --        & --        \\
Large       & 307M / 24 & 307M / 24 & 307M / 24 & 300M / 24 & 307M / 24 & 300M / 24 & 307M / 24 & 307M / 24 \\
Huge        & 632M / 32 & --        & --        & 840M / 32 & 632M / 32 & 600M / 24 & --        & --        \\
Giant/1B    & --        & --        & 1.1B / 40 & --        & --        & $\sim$1B / 24 & --    & --        \\
\bottomrule
\end{tabular}
}
\caption{Models used in the cross-paradigm (Fig.~\ref{fig:cross_model}) and
cross-scale (Fig.~\ref{fig:cross_scale}) experiments, by scale tier, with
each family's pretraining paradigm noted beneath its name. Cells give
parameter count / block count; a dash means no public checkpoint at that
tier. The six Large-tier models are also the backbones used in
App.~\ref{sec:dt1}; the fine-tuning-depth sweep (App.~\ref{app:lora-setup})
uses a different six-backbone set spanning both scale and paradigm,
detailed there. We compare models at
matched relative depth, not block index (Sec.~\ref{sec:method:probe}), so
differing block counts per row are not a confound -- except AIMv2, which
fixes 24 blocks at every tier and scales width instead, so its Huge/Giant
cells are not block-matched to the rest of that row.}
\label{tab:model_registry}
\end{table}

\begin{table*}[t]
  \centering
  \footnotesize
  \setlength{\tabcolsep}{4pt}
  \begin{tabular}{llllll}
    \toprule
    Task & Dataset & Decoder head & Train $n$ & Epochs & Metric \\
    \midrule
    Classification         & Tiny-ImageNet-200      & Pool + linear              & 100{,}000 & 150 & Top-1 acc. \\
    Detection               & COCO val2017            & CenterNet                  & 4{,}457   & 145 & mAP@50 \\
    Semantic segmentation   & ADE20K                  & Single $1{\times}1$ conv   & 20{,}210  & 100 & mIoU \\
    Instance segmentation   & COCO val2017 (instance) & CenterNet + mask conv      & 4{,}457   & 145 & Mask mAP@50 \\
    Panoptic segmentation   & COCO-Panoptic           & Single $1{\times}1$ conv   & 4{,}500   & 143 & PQ-proxy mIoU \\
    Depth                   & NYU Depth V2            & Single $1{\times}1$ conv   & 1{,}305   & 250 & $\delta_1$ \\
    Surface normals         & NYU Depth V2 (derived)  & Single $1{\times}1$ conv   & 1{,}305   & 250 & Acc.\ @11.25$^\circ$ \\
    Pose                    & COCO Keypoints          & $K$-channel conv           & 5{,}717   & 112 & PCK@0.5 \\
    Retrieval                & GLDv2 (mini-subset)    & Pool + linear + $\ell_2$   & $\gtrsim$12{,}736 & 50 & Recall@1 \\
    \bottomrule
  \end{tabular}
  \caption{The 9 tasks behind the frozen-backbone benchmark: dataset,
  decoder head, training-set size, and epoch budget for each (protocol in
  App.~\ref{sec:dt1}; results in Table~\ref{tab:benchmark_results}, pooled
  score in Fig.~\ref{fig:benchmark}). These are the `9 popular, divergently
  specialized tasks' mentioned in the main text, chosen to share no
  architectural structure so a backbone's pooled score reflects general
  representation quality rather than skill at one task family.}
  \label{tab:benchmark}
\end{table*}

\subsection{Fine-tuning sweep: full setup} \label{app:lora-setup}
\begin{figure}[ht]
    \centering
    \includegraphics[width=0.6\textwidth]{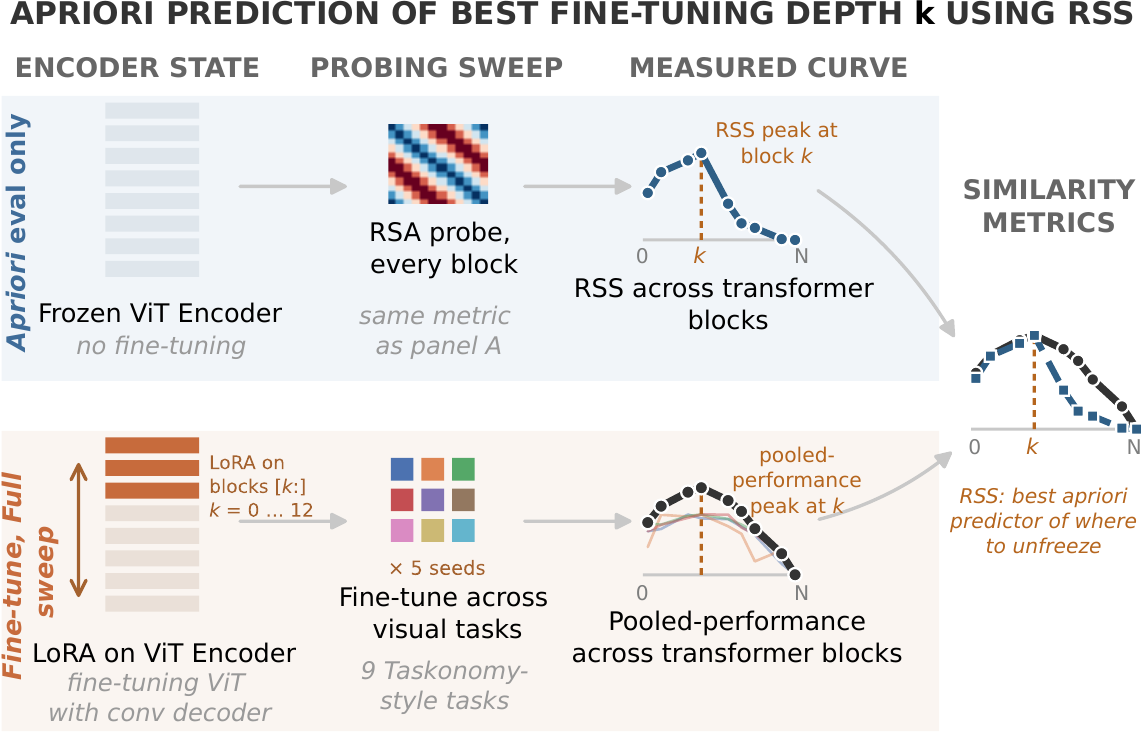}
    \caption{Protocol schematic. \emph{Top:} the frozen backbone is probed
    once, giving an a-priori RSS-vs-block curve. \emph{Bottom:} the same
    backbone is adapted with LoRA on blocks $[k:]$ and fine-tuned across
    the 9 Taskonomy-style tasks (5 seeds), giving an empirical
    performance-vs-$k$ curve. RSS's peak coincides with the performance
    peak, making it an a-priori predictor of where to unfreeze.}
    \label{fig:downstream}
\end{figure}

\paragraph{What this experiment tests} This is the fine-tuning-depth sweep
behind Fig.~\ref{fig:fine-tuning} and App.~\ref{app:apriori-ranking},
schematized in Fig.~\ref{fig:downstream}. For each of six backbones, we
adapt progressively more of the network with LoRA from a range of starting
depths $k$, measure downstream performance as a function of $k$, and
compare it to RSS's a-priori, no-fine-tuning-required curve on the same
frozen backbone -- testing whether RSS's peak predicts where fine-tuning
performance actually peaks. These six backbones are chosen to test whether
that prediction holds both across scale and across paradigm, not just for
one representative model: DINOv3 at the S, B, and L tiers tests scale,
since DINOv3 is the best-performing backbone in the frozen-backbone
benchmark (App.~\ref{sec:dt1}); MAE, AIMv2, and DeiT3 test paradigm,
chosen as the frozen-backbone benchmark's most divergent RSS shape (MAE),
worst performer (AIMv2), and a mid-range performer (DeiT3),
respectively.

\paragraph{Depth grid and LoRA config} For a backbone with $L$ blocks
($0,\dots,L-1$), we unfreeze from block $k$ onward over nine relative
depths $d=k/(L-1)$, fixed once on DINOv3 ViT-S/16 ($L=12$):
$k\in\{0,1,3,4,6,7,8,10,11\}$, $d\in\{0,.09,.27,.36,.55,.64,.73,.91,1\}$.
Every other backbone adapts from the same nine relative depths (nearest
block index), so results stay comparable despite differing depth. Blocks
before $k$ stay frozen; blocks from $k$ on are adapted with LoRA
\citep{hu2022lora}, which adds a trainable low-rank residual to a frozen
linear layer $W$:
\begin{equation}
  W'x = Wx + \frac{\alpha}{r}(BA)x, \label{eq:lora}
\end{equation}
with $A,B$ trainable rank-$r$ factors, $\alpha$ a fixed scale, and $B$
zero-initialized so training starts at the pretrained function. We use
$r=8,\alpha=16$ on every linear layer (attention QKV/projection, MLP) in
blocks $[k:]$ -- $\mathrm{LoRA}(\text{dino},k)$ -- with the same
task-specific decoder heads as the frozen-backbone benchmark
(Sec.~\ref{sec:dt1}) on top, adding no capacity to compensate for what the
backbone lacks. Each task also gets one fully frozen, decoder-only baseline
(no LoRA), so 10 configurations per task in total. Locating each
backbone's transformer blocks and adding LoRA adapters to their linear
layers is implemented generically, not hardcoded to one model family, so
the same pipeline runs unmodified on any ViT backbone used in this paper.
The six backbones actually swept here are listed in the main text.

\paragraph{Optimization} We optimize with AdamW (weight decay $10^{-4}$)
at a constant learning rate of $3\times10^{-4}$ for both the LoRA adapters
and the decoder head, with no learning-rate schedule. Images are resized
to $224\times224$ for every backbone, keeping the input pixel budget fixed
across models and depths so the LoRA start-layer comparison is not
confounded by resolution differences. Batch size is tiered to GPU memory:
16 for LoRA-adapted runs and 24 for the frozen baseline, at the L tier. We
train for 300
epochs per (task, $k$, seed), evaluate every 25 epochs, and keep the best
checkpoint -- as in Sec.~\ref{sec:dt1}, this is best-checkpoint selection
over a fixed budget, not early stopping.

\paragraph{Dataset} We fine-tune on a subset of Taskonomy-tiny
\citep{zamir2018taskonomy}, a widely used benchmark for generalization
across varied vision tasks: 8 buildings (hanson, merom, klickitat, onaga,
leonardo, marstons, newfields, pinesdale), up to 150 RGB frames each,
aligned across all 9 task modalities to 1140 usable frames, split 983
train / 157 val (reshuffled per seed). This is the largest split we can
afford at this sweep's scale -- $9$ tasks $\times$ ($9$ LoRA depths
$+\,1$ frozen) $\times$ $5$ seeds $=450$ runs per backbone -- while still
large enough that real differences in performance across unfreeze depths
$k$ are distinguishable from ordinary run-to-run variation across seeds.
We average
each (task, $k$) cell over 5 seeds and report mean min-max normalized
performance, reusing Eq.~\eqref{eq:mean-norm-perf} with $k$ in place of
$b$; unlike Sec.~\ref{sec:dt1}'s cross-backbone pooling, $b'$ here ranges
only over one backbone's own depths, normalized independently per
backbone, since this sweep asks where to unfreeze a chosen backbone, not
which backbone is best (seed spread shown separately in
Fig.~\ref{fig:fine-tuning}). Because Taskonomy labels many modalities of
the same images, it lets us test several genuinely different tasks on one
standardized dataset.

\paragraph{Tasks} We derive 9 tasks from Taskonomy's per-pixel modalities,
chosen as its most commonly benchmarked tasks and varied enough to test
generalizability across whole-image to dense-pixel encoding: semantic
segmentation (mIoU); depth and Euclidean distance ($\delta_1$
accuracy); detection, boxes derived from the segmentation masks (mAP@50);
surface normals (accuracy within $11.25^\circ$); reshading (PSNR); and
principal curvature, edges, and 2D keypoints (mean absolute error). These
span genuinely different structure, not relabelings of the same one:
segmentation/detection are semantic, depth/Euclidean distance/reshading
depend on 3D scene geometry and lighting, normals/curvature depend on local
surface geometry, and edges/keypoints are comparatively low-level. A
backbone that transfers well to only one family would not support a
general-purpose claim; scoring all nine together is what makes the pooled
performance in Eq.~\ref{eq:mean-norm-perf} a meaningful generalization
proxy rather than a result specific to one task's idiosyncrasies.

\paragraph{Compute} Every (task, $k$, seed) run trains independently, with
several runs sharing one GPU (batch size and concurrency tuned to the
backbone's size). All runs used a single NVIDIA H200; the full 450-run
sweep for DINOv3 ViT-L/16 took about 3.5 days of wall-clock time. We
repeat this protocol across the six backbones described above: DINOv3
ViT-S/16, ViT-B/16, and ViT-L/16 (scale), and MAE, AIMv2, and DeiT3 at the
L tier (paradigm).

\subsection{Ranking RSS among 9 a-priori predictors} \label{app:apriori-ranking}
This section asks whether RSS's ability to predict the best fine-tuning
depth (App.~\ref{app:lora-setup}) is a genuine signal or just a lucky
choice of metric. We score RSS alongside 9 a-priori predictors computed
from the frozen backbone alone, before any fine-tuning: 5 orientation
metrics (RSS, recruitment, vector OSI, linear decoding, von Mises
bandwidth) and 4 equivariance metrics (horizontal flip, $90^\circ$
rotation, translation, their mean), each at block $k$ only, $5+4=9$
predictors total. We drop each metric's mean over $[k:]$: a running mean
mixes signal from blocks the fine-tuning run hasn't reached yet with ones
it has, so it isn't a meaningful a-priori predictor. Each predictor is
scored by correlation with performance across $k$ (Pearson/Spearman), a
composite shape-similarity score (mean of min--max overlay, shift-tolerant
DTW, and top-half overlap), and argopt-$\Delta$ (gap between the
predictor's and performance's peak block). RSS ranks first of 9 on every
measure: shape similarity $0.76$, $|$Pearson$|=0.83$, $|$Spearman$|=0.92$,
argopt-$\Delta=0$ ($k=4$ exactly). The same holds scoring per-task rather
than pooled (shape $0.72$, pooled $|$Spearman$|=0.60$, still first). No
single statistic here is a formal significance test; what matters is that
three qualitatively different measures agree, and agree specifically for
RSS, out of 9 task-agnostic candidates computed before any fine-tuning.
\end{document}